\pdfoutput=1
\documentclass[letterpaper, 10 pt, conference]{ieeeconf}  

\IEEEoverridecommandlockouts                              
\usepackage[pdftex]{graphicx}
\DeclareGraphicsExtensions{.pdf,.jpeg,.png}
\usepackage{amsmath} 
\usepackage{url}
\usepackage{subfigure}
\usepackage{color}
\definecolor{darkgreen}{RGB}{0,100,0}
\definecolor{gray}{rgb}{0.5,0.5,0.5}

\usepackage{listings}

\title{\LARGE \bf
Dealing with Run-Time Variability in Service Robotics:\\
Towards a DSL for Non-Functional Properties
}

\author{Juan F. Ingl{\'e}s-Romero$^{1}$, Alex Lotz$^{2}$, Cristina Vicente-Chicote$^{1}$ and Christian Schlegel$^{2}$
\thanks{$^{1}$J.F. Ingl{\'e}s-Romero and C. Vicente-Chicote are with ETSIT, Tech. Univ. of Cartagena, Spain,
        {\tt\small \{juanfran.ingles, cristina.vicente\} at upct.es}}%
\thanks{$^{2}$A. Lotz and C. Schlegel are with the Computer Science Department, University of Applied Sciences Ulm, Germany
        {\tt\small \{lotz, schlegel\} at hs-ulm.de}}%
}

\begin{document}

\maketitle
\thispagestyle{empty}
\pagestyle{empty}

\begin{abstract} 

 Service robots act in open-ended, natural environments. Therefore, due to 
 combinatorial explosion of potential situations, it is not possible to foresee
 all eventualities in advance during robot design. In addition, due to limited resources on a 
 mobile robot, it is not feasible to plan any action on demand. Hence, it is 
 necessary to provide a mechanism to express variability at design-time 
 that can be efficiently resolved on the robot at run-time based on the 
 then available information. 
 
In this paper, we introduce a DSL 
to express run- time variability focused on the execution quality of the
robot (in terms of non-functional properties like safety and task efficiency) 
under changing situations and limited resources. We underpin the
applicability of our approach by an example integrated into an overall robotics
architecture.

\end{abstract}

\section{Introduction}

%
%

Advanced robotic systems like service robots (robot companion, elder care, home health care, 
robot co-worker) are expected to robustly and efficiently fulfill
different tasks (multi-purpose systems) in complex environments (domestic, outdoor, public spaces). While robots are always only 
equipped with a limited set of resources (processing power, energy supply, sensing capabilities,
skills), real-world environments are inherently open-ended and show a huge
number of variants and contingencies. Thus, for many different situations, a
robot needs to know how to spend its scarce resources in the most appropriate way (in short, acting
efficiently) in order to achieve a high degree of robustness and to maintain a high success rate in task 
fulfillment. Although to date the focus still is mostly on pure task achievement
(i.e., on robot functionality like in \cite{Boren:2010}), balancing
non-functional properties (e.g., quality of service, safety and energy
consumption) becomes more and more important and cannot be ignored in advanced service robotic systems.

Even the most skilled robotics engineer is not able to identify and enumerate all eventualities 
in advance and to properly code configurations, resource assignments and reactions at design-time
(not to mention that this is not efficient at all due to the combinatorial explosion of possible 
situations and skill parameterizations). Unfortunately, it is also not possible just to \mbox{(re)plan} 
at run-time in order to take into account latest information as soon as it becomes available. The
computational complexity of planning is far too high when it comes to real-world problems (i.e., 
generate action plots given partial information while taking into account additional properties 
like safety and resource awareness).

This motivates a different approach: \mbox{(i) we} need to make it as simple as possible for 
the designer to {\em express variability at design-time} and \mbox{(ii) we} need the robot to 
be able to {\em bind variability at run-time} based on the then available information. At design
time, we also specify which problem solver (symbolic planner, constraint solver, etc.) to use to 
bind which variation point \cite{Schlegel:2010}, \cite{Steck:2011}. At run-time, the robot then 
involves the prearranged and dedicated problem solvers. Overall, this improves task execution 
quality, optimizes robot performance and cleverly arranges complexity and
efforts between design-time and run-time.

\begin{figure}[thpb]
  \centering
  \includegraphics[width=\columnwidth]{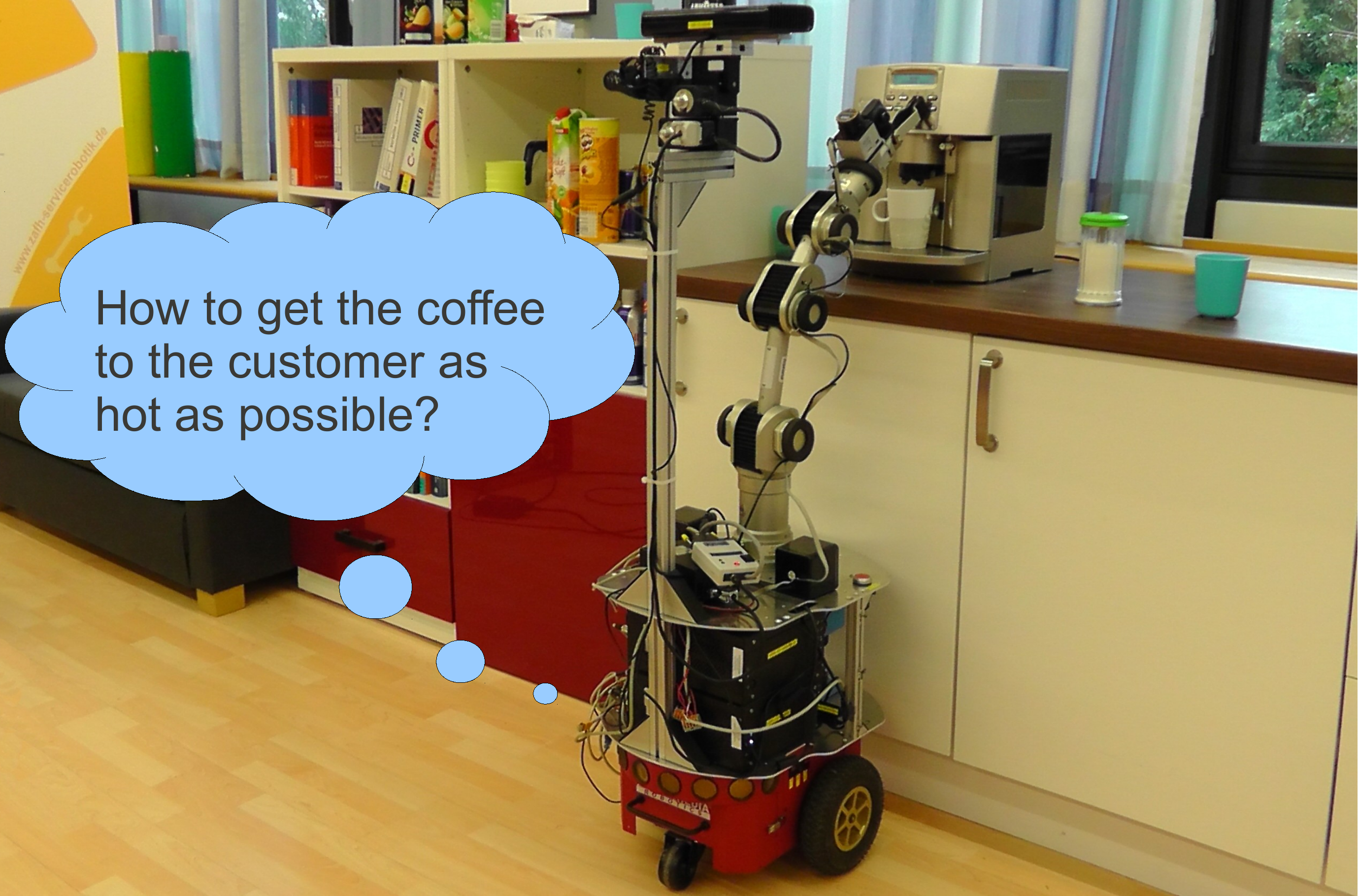}
  \caption{Our robot Kate makes coffee}
  \label{fig:chs-scenario}
\end{figure}

Let's illustrate this by an example scenario: a service robot delivers coffee on demand (see 
fig. \ref{fig:chs-scenario}). In order to optimize the coffee delivery service, the robot has 
to trade-off various aspects to come up with an optimized quality of service. The robot needs
to be able to select an appropriate velocity (variation point) to
properly fulfill its task according to further issues like safety and energy:
\mbox{(i) customers} are satisfied only if the coffee has at least a certain temperature but 
    prefer it as hot as possible,
\mbox{(ii) however,} the maximum allowed velocity is bound due to safety issues (hot coffee)
    and also by battery level,
\mbox{(iii) since} coffee cools down depending on the time travelled, a minimum
required average velocity (depending on distance to customer) is needed, although driving slowly
    makes sense in order to save energy,
\mbox{(iv) nevertheless}, fast delivery can increase volume of coffee sales.

The late binding
of variability (variation points purposefully left
open by the designer) allows for a clever way of complexity handling according
to the following principles:

{\em Separation of concerns:} a model (e.g., a task net) describes {\em how} to deliver a coffee
(i.e., the action plot) \cite{Steck:2011}. Further models specify {\em what is a good way} (policy) of delivering 
a coffee (e.g., in terms of non-functional properties like safety or energy
consumption).

{\em Separation of roles:} the {\em designer} provides at design-time the models (action plots 
with variation points to be bound later by the robot, policies for task
fulfillment, problem solvers to use for binding of variability). The {\em robot} decides at run-time on proper bindings 
for variation points by applying the policies and taking into account current
situation and context.

With this approach, the design-time modeling effort stays feasible even when extending task 
plots to non-functional properties like safety and energy consumption. On the other hand, 
even under non-functional constraints, run-time decisions on variation points become feasible
due to narrowed search spaces.
At the same time, this allows to much better
address open-ended environments since policies (what is a good way of parameterizing a task
plot) can come up at run-time with bindings of variability without enumerating all possible
situations in advance. However, this requires means to express variability at design-time 
-- in particular on non-functional properties -- 
such that it can be exploited at \mbox{run-time}. 

In this paper, we introduce a first version of a Domain-Specific Language (DSL)
for expressing run-time variability.
It provides a mechanism to specify how a system should adapt to cope with changing situations 
in order to maintain or improve the execution quality of the system (in terms of non-functional properties 
like safety and task efficiency). We also describe the integration of the DSL into an overall robotics architecture. 
A real-world example underpins the feasibility of the approach and illustrates its benefits. We conclude with 
hints on next steps towards a full-fledged DSL and system integration.

\section{Variability Modeling Language}
\label{sec:method}

%
%

In this section we present the \emph{Variability Modeling Language (VML)}
that provides a mechanism to express how a system should adapt at run-time based
on non-functional properties and adaptation rules. First, we introduce the
overall idea behind the proposed language. After that, we address the VML syntax
and the specification of the semantics.

\subsection{Overview}

Following the coffee-delivering example presented in the introduction, the
service quality of a system can be given in terms of non-functional properties,
such as, safety, performance and energy consumption. These properties are often
contradictory and conflicting, and their importance varies according to the
current system context (e.g., if the battery of the robot is low, energy
consumption has a higher priority). Properties are expressed as functions of
variation points, i.e., system variables left open at design time and bound at
run-time (e.g., the average velocity impacts on performance since a fast
delivery can increase the volume of coffee sales). Therefore, binding variation
points results in a certain quality of service (e.g., depending on the concrete
value for the average velocity, the robot is more or less efficient, safe or
energy consuming). The objective is to maximize the quality of service through
the configuration of the variation points each time the system context changes
significantly. Finally, the variation points can be constrained depending on
the context using adaptation rules. As shown in figure 2, we have already
presented all the keys of VML: \emph{variation points}, \emph{context},
\emph{properties} and \emph{adaptation rules}.

\begin{figure}[thpb]
  \centering
  \includegraphics[width=0.9\columnwidth]{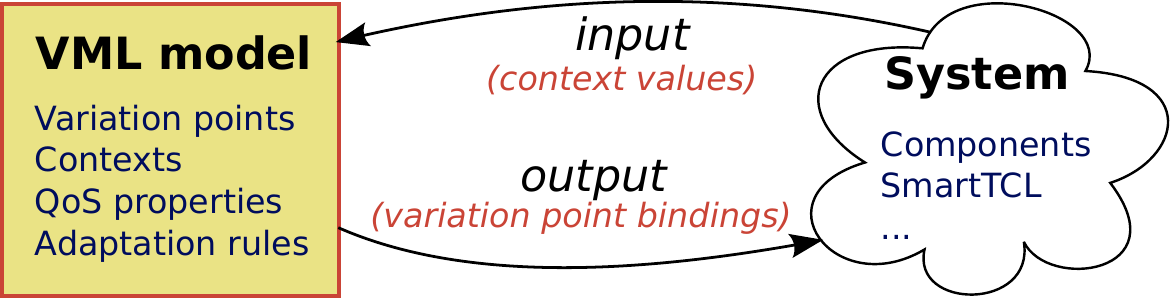}
  \caption{VML Models and the interaction with the system (see sec. \ref{sec:integration-into-robotics})}
  \label{fig:vml-introduction}
\end{figure}

\subsection{VML Syntax}

Listing \ref{lst:ebnf} shows the EBNF grammar of the VML language. The use
of VML will be later illustrated in section \ref{sec:realworldexample} with
full-fledged examples.

\lstset{ %
  basicstyle=\tiny\ttfamily,           
  numbers=none,                   
  numberstyle=\tiny\color{gray},  
  stepnumber=1,                   
  mathescape=false,
  numbersep=5pt,                  
  backgroundcolor=\color{white},      
  showspaces=false,               
  showstringspaces=false,         
  showtabs=false,                 
  frame=single,                   
  rulecolor=\color{black},        
  tabsize=2,                      
  captionpos=b,                   
  breaklines=true,                
  breakatwhitespace=false,        
  title=\lstname,                   
  keywordstyle=\color{black},          
  commentstyle=\tiny\color{gray},       
  stringstyle=\color{blue}\bfseries,         
  morestring=[b]',
  morekeywords={ID,REAL,INT}               
}

\begin{lstlisting}[caption={EBNF grammar of the Variability Modeling Language
(VML). Due to lack of space, we only include a reduced version of the grammar
instead of a full syntax specification.}, label=lst:ebnf] 
VMLModel ::= ( TypeDefinition | VariableDefinition | AdaptationRule )+
TypeDefinition ::= EnumType | NumericType | BooleanType 
EnumType ::= 'enum' ID { ( EnumLiteral ; )+ }
EnumLiteral ::= ID ('(' INT ')')?
NumericType ::= 'number' ID '{' 'range' ':' '[' (REAL|INT) ',' (REAL|INT) ']' ';' 
  'precision' ':' (REAL|INT) ';' ( 'unit' ':' STRING ';' )? '}'
BooleanType ::= 'boolean' ID ';' 
VariableDefinition ::= GeneralVar | Context | VariationPoint | Property
GeneralVar ::= 'var' ID ':' ID '=' expr ';'
Context ::= 'context' ID ':' ID ';'
VariationPoint ::= 'varpoint' ID ':' ID ( ';' | '{'
  (InvariantDefinition | ImplicationDefinition) (',' 
  (InvariantDefinition | ImplicationDefinition) )* ';' '}' 
Property ::= 'property' ID ':' ID ('maximized'|'minimized')? '{'
  'priorities' ':' FunctionDefinition (',' FunctionDefinition)* ';'
  'definitions' ':' FunctionDefinition ( ',' FunctionDefinition)* ';' '}'
AdaptationRule ::= 'rule' ID ':' ImplicationDefinition';'
\end{lstlisting}

VML is a declarative language that allows the definition of: data types,
variables, and adaptation rules. Data types are used to define the nature of the
variables, i.e., their possible values and operations. VML includes three basic
data types: (i) enumerators; (ii) ranges of numbers; and (iii) booleans.
\emph{Enumerators} are defined by a list of literals that represent a set of
unique values masked by alias. Per default, the $i$-th literal in the enumerator
is coded as the integer $i$. \emph{Ranges of numbers} define intervals of
discrete values. They require the specification of the interval limits and the
precision to discretize the interval. Optionally, they can declare the physical
unit (e.g., m/s for velocity), which is taken into account for conversions and
normalization (further details in next subsection). Arithmetic operations can be
applied to enumerators and ranges of numbers while logical operations to
booleans.

Regarding variables, VML includes three kinds of special-purpose variables
(\emph{Contexts}, \emph{Variation points} and \emph{Properties}) and one
general-purpose variable for auxiliary calculations. Each variable is declared
to belong to a certain data type, which needs to have been previously defined
as explained before. \emph{Context variables} specify the internal and external
environment features on which adaptation depends (e.g., listing
\ref{lst:vml-velocity} / section \ref{sec:realworldexample}:
\texttt{ctx\_battery} is a context variable).
\emph{Variation points} denote the variability of the system (e.g., listing
\ref{lst:vml-velocity}: \texttt{maximumVelocity}). The definition of
these variables may optionally include a set of invariants and implications
(used to constrain their possible values and to define their dependencies on
other variation points).

At this point, we have introduced context variables, which are intended to
capture the system context, and variation points, which represent the
variability that is left open at design-time and must be bound at run-time. In
this sense, we can consider context variables as inputs of the adaptation
process and (the values selected for all) variation points as its outputs. Now,
we introduce the elements that describe how the system should adapt, i.e., how
the outputs must be set considering the inputs. This is achieved through
\emph{properties} (the last special-purpose kind of variable), and
\emph{adaptation rules}. Properties specify the features of the system that
need to be optimized, i.e., minimized or maximized (see
\texttt{energyConsumption} and \texttt{performance} in listing \ref{lst:vml-velocity}).
Properties are defined by two kinds of functions: \emph{priorities} and \emph{definitions}.
\emph{Definitions} describe the property in terms of one or more variation
points; e.g., in listing \ref{lst:vml-velocity}, the \texttt{energyConsumption}
property is defined as an exponential function depending on the
\texttt{maxVelocity} variation point, meaning that changes in the maximum speed
have an exponential impact on the energy consumption. On the other hand,
\emph{priorities} describe the importance of each property in terms of one or
more context variables; e.g., \texttt{energyConsumption} becomes more and more
relevant as the robot battery (\texttt{ctx\_battery}) decreases, resp., when the
battery is full, energy consumption is not considered an issue and, thus, its impact on the
adaptation process is very small. Note that the definition of these functions
has been obtained empirically.

\emph{Adaptation rules} define direct relationships between the context
variables and the variation points (properties also relate them but
indirectly). \emph{Adaptation rules} are \emph{ECA (event-condition-action)}
rules, i.e., their left-hand side expresses a condition (depending on one or
more context variables) for the rule to be triggered, and their right-hand side
constrains the output variation point (i.e., the values it can take). 
For example, the rule \texttt{low\_noise} in listing \ref{lst:vml-velocity}
forces \texttt{speakerVolume} to 35 when \texttt{ctx\_noise} is less than 20. 

\subsection{VML Execution Semantics and Implementation}

Executing a VML model means finding the best configuration possible (bindings
for the variation points) given the current context. Let $ctx$ be an $n$-tuple
of the values associated to each context variable in the model, i.e,
$ctx=c_0,\dots,c_{n-1}$ where $c_i$ is the value of the $i$-th context variable
and $n$ the number of context variables in the model. Let $vp$ be an $m$-tuple
of the values associated to each variation point in the model, i.e.,
$vp=v_0,\dots,v_{m-1}$ where $v_i$ is the value of the $i$-th variation point
and $m$ the number of variation points in the model.

Given a certain context $ctx$, we need to find $vp$ that minimizes the cost
function $f(ctx,vp)$ subject to the constraints imposed by the
adaptation rules that verify the right side condition and by the dependencies
of the variation points:

\begin{equation}
\label{eq:costfkt}
f(ctx,vp) = \sum_{\forall i} (-1)^{d_i} \cdot w_i(ctx) \cdot p_i(vp)
\end{equation}

$w_i(ctx)$ is the normalized sum of the priority functions, $p_i(vp)$ the
normalized sum of the definition functions associated to the $i$-th property in
the model, and $d_i$ specifies whether the $i$-th property should be minimized
($d_i = 0$) or maximized ($d_i = 1$).

The normalization process is required in order to make variables comparable.
First, the process homogenizes the variable units (if declared), and then, the
variables are scaled to $[0,100]$ according to their range definitions in
the models.

For execution of VML models, we took advantage of existing constraint solvers to
deal with the required constraint optimization problems. We have selected
\emph{MiniZinc}~\cite{Nethercote:2007} for the run-time support of the VML models, since it is a simple and
expressive modeling language, which is independent of the solver.
\emph{MiniZinc} is currently supported by many constraint programming systems, among others, the \emph{G12
Constraint Programming
Platform}\footnote{\url{http://www.g12.csse.unimelb.edu.au/}}. 

Although this is work in progress, we have created a preliminary textual
editor for the VML language using the \emph{Model-Driven Engineering (MDE)}
tools available in Eclipse~\cite{Gronback:2009}. The editor has some advanced
features such as syntax checking, colouring and completion.

\section{Integration Into A Robotics Architecture}
\label{sec:integration-into-robotics}

%
%

\begin{figure*}[thpb]
  \centering
  \includegraphics[width=\linewidth]{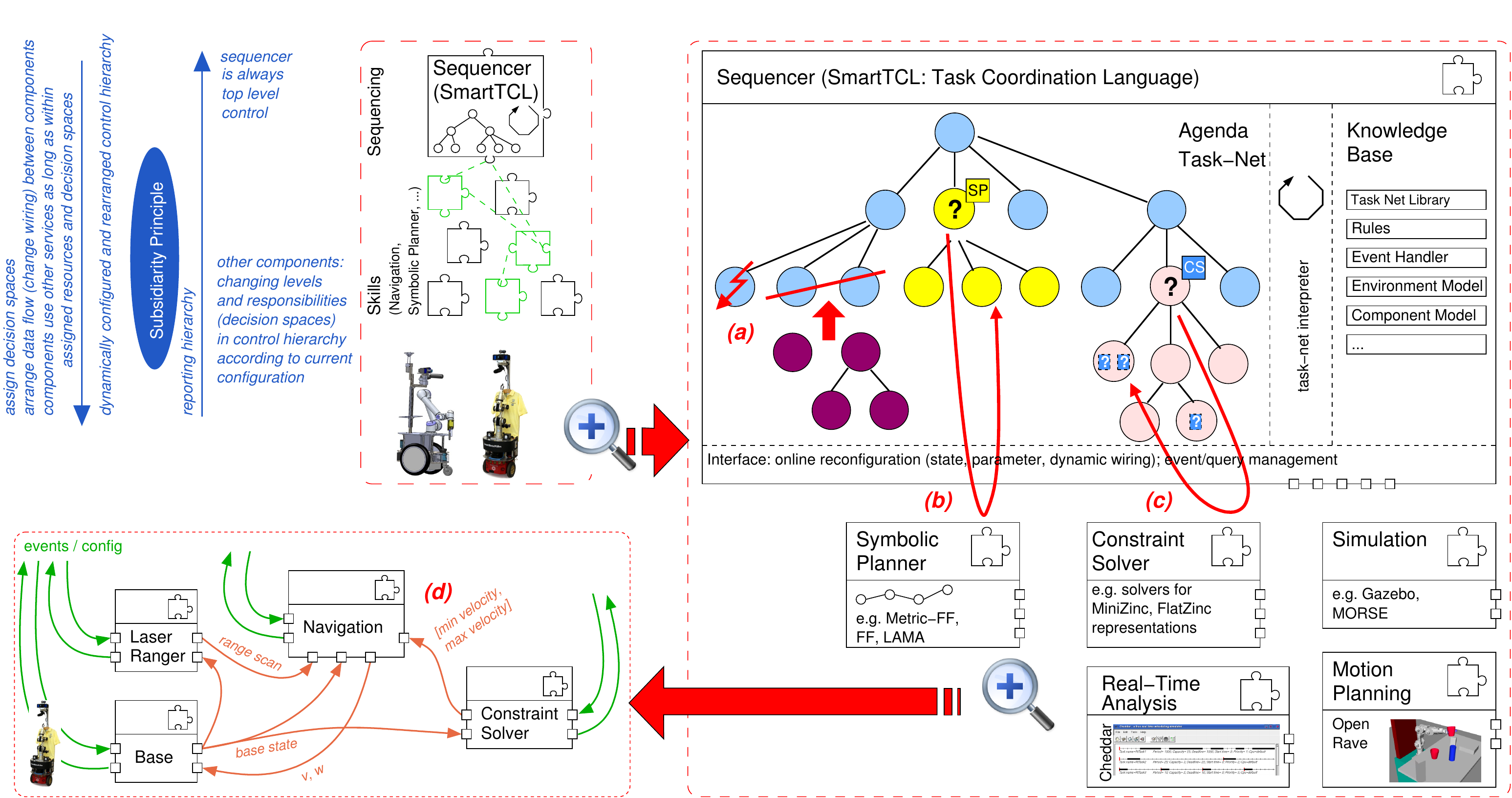}
  \caption{Exploiting variability in an integrated robotics architecture. In {\sc SmartTCL},
    a task-tree is specified at design-time. Several variation points are left open and bound
    at run-time:
    \mbox{(a) reacting} to execution flaws by adapting task-net expansions to
    the current situation, \mbox{(b) expert} called to give advice (query): symbolic planner as 
      expert for situation-dependent run-time task node expansion (e.g., 
      planning how to stack the different cups into each other when cleaning 
      a table),
    \mbox{(c) expert} called to give advice (query): a constraint solver as 
      expert for run-time binding of variation points in task nets (e.g., binding 
      maximum velocity depending on safety constraints according to current payload),
    \mbox{(d) expert} configured for periodic advice (push): configure for advice on
      orthogonal parameters (e.g., constraining minimum / maximum velocity as decision
      space for navigation component as in coffee delivery example; the navigation 
      then decides within given changing boundaries on actual velocity taking into
      account current obstacles).
  }
  \label{fig:chs-architecture}
\end{figure*}

The overall integration into a robotics architecture is shown in figure
\ref{fig:chs-architecture}. The {\em sequencer} orchestrates the system and
its software components \cite{Steck:2011} (send parameters/configurations;
switch components on/off; change the wiring between components; query information; wait for
events; coordinate analysis, simulation and planning capabilities).

The overall approach of task decomposition for behavioural variability follows 
the {\em subsidiarity principle} in order to cope with the challenge of incorporating
different control schemes. The principle of the run-time task execution can be compared
with a company where budgets and responsibilities are assigned down the hierarchy and
where, at the same time, rational decisions and activities are expected.
Hierarchy is needed in order to ensure that the assigned decision spaces are not in conflict with
each other (these have to be either orthogonal or exploit only a limited budget of 
a shared resource). Hierarchy is motivated by the fact that a single robot, as
one entity, can at the end only be at one location. Indeed, hierarchy is
compatible with decentralized control (including adaptation capabilities at
different hierarchical layers) promoting autonomous or semi-autonomous
adaptation to rapidly changing environments and circumstances (which is essential in robotics).

Thus, following the subsidiarity principle, we have changing hierarchies with 
decentralized control following a delegation model. Task execution results in task
refinements which come along with run-time binding of purposefully left open variability. 
Coming up with more and more bindings down the hierarchy of task refinements imposes 
constraints onto the subordinated entities (or put it the other way round: assigns 
down the hierarchy decision spaces as broad as possible while ensuring that these
locally exploitable variability is not in conflict with concurrent activities).

The sequencer plays the master role in our multilayered robotic architecture and
assigns decision spaces down the hierarchy to the skill components (see fig.
\ref{fig:chs-architecture} on the top left). Thereto, the sequencer parametrizes
skill components such that they can bind variation points within predefined
boundaries. For example, according to the current task, the sequencer sets
reasonable boundaries for maximum translation and rotation velocities in the
\emph{Curvature Distance Lookup (CDL)}~\cite{Schlegel:1998} component for
obstacle avoidance. CDL is then free to choose slower velocity values (e.g., in
case an obstacle is in the way of the robot). Thus, the sequencer delegates the
responsibility to bind the variation point \texttt{maximumVelocity}
(list.~\ref{lst:vml-velocity}) within predefined limits to the CDL component.
Second, the sequencer itself is able to deal with variability as illustrated in
(a) and (b) (fig. \ref{fig:chs-architecture}). In (a) it is possible to model
sophisticated recovery strategies (e.g., in case a skill component fails to
fulfill its task or the situation resp. context around the robot changes). In
(b) it is also possible to purposefully leave open the expansion plan for
certain nodes of the task-net at design-time to ask a symbolic planer (as an
expert for certain problems) later at run-time for a proper sequence of
child-nodes. The cases (c) and (d) are of particular importance since they 
show the interactions between the sequencer, the skill components and the
constraint solver which physically runs the VML models (see next
section for more details).

\section{Full-Fledged Real-World Example}
\label{sec:realworldexample}

\begin{figure}[htpb]
   \centering
   \includegraphics[width=1.0\columnwidth]{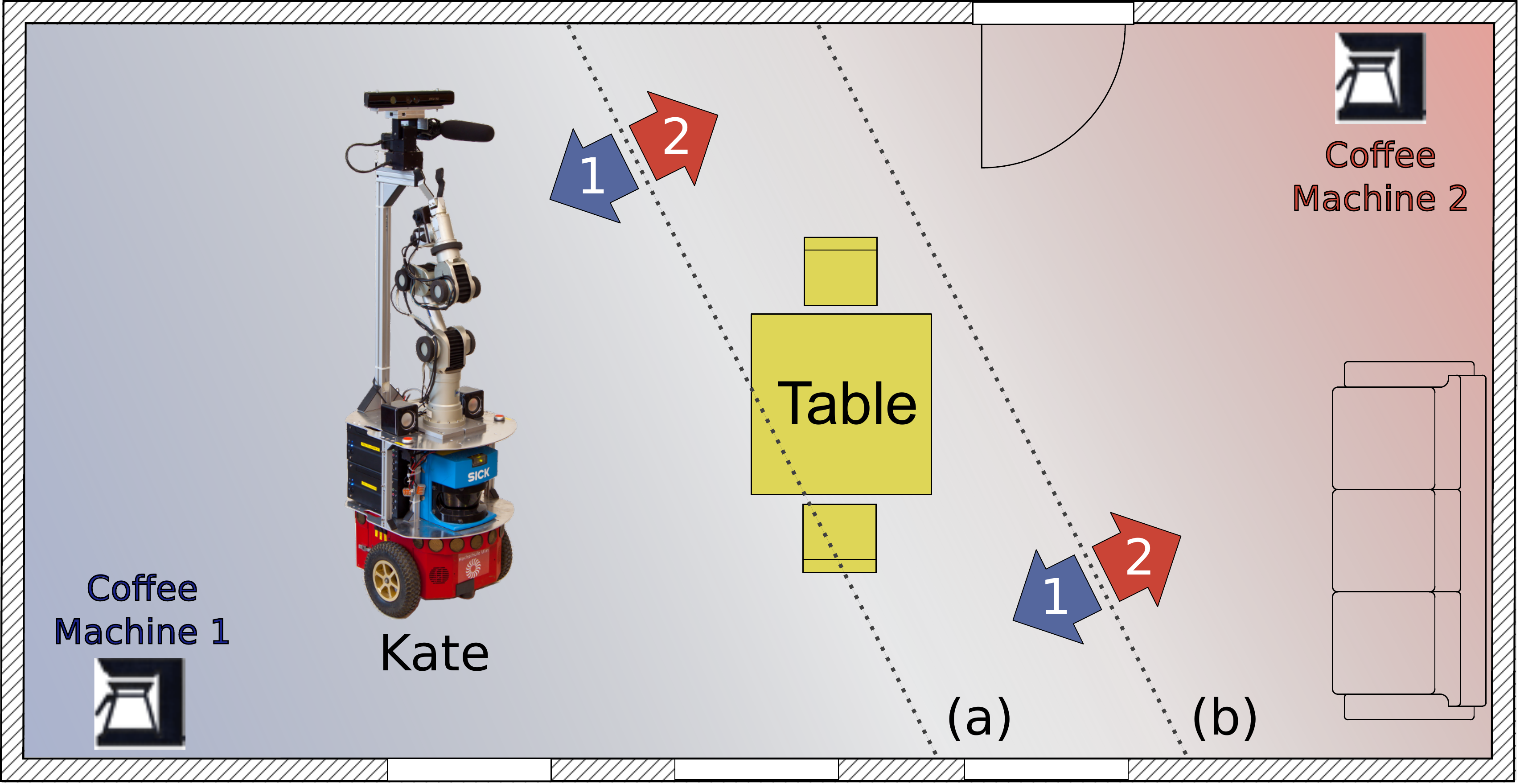}
   \caption{Depending on the current distance between Kate and the coffee machines, the waiting time at the coffee machines and Kate's current battery
   level, the decision threshold to drive to coffee machine 1 or 2 moves either towards (a) or (b).}
   \label{fig:scenario}
\end{figure}

%
%
In this section, we illustrate our proposal by a simple scenario based on the
coffee-delivering example given in the introduction. The case study takes place
in a room with two coffee machines located in different positions (see
fig.~\ref{fig:scenario}). Our robot called Kate, moves around in the scenario
meeting people. Thus, when someone asks her for a cup of coffee she must decide:
(i) which coffee machine she will use, (ii) her maximum allowed velocity, and
(iii) her speaker volume (speak up in noisy environments while not shouting in
quiet environments significantly improves social acceptance of the robot).
This decision is made at run-time in order to improve the quality of the service taking into account \texttt{energyConsumption} (e.g.,
when the battery is low the system must optimize \texttt{energyConsumption} using the nearest coffee
machine) and \texttt{performance} (e.g., trying to get the highest value for maximum
allowed velocity in order to reach the goal earlier). 
Obviously, maximizing \texttt{performance}, while simultaneously minimizing
\texttt{energyConsumption}, imposes conflicting requirements. Thus, the adaptation
strategy will need to find the right balance among these requirements to
achieve the best possible configuration for a given context, even if some (or
none) of them are optimized individually.

The scenario has been specified in two VML models. Listing \ref{lst:vml-coffee}
shows the first one, which includes the following context variables: (i) the
\texttt{ctx\_battery} (integer value in the range 5-100); (ii) distance to each
coffee machine (real number in the range 0-20 with precision 0.1 and meters); (iii) waiting
time at each coffee machine (integer in the range 10-300 and seconds), it
considers the operation time of the machine (constant time) and the time that
the robot has to wait because the machine is busy and there may be others
(robots or people) waiting to use the machine (variable time); and (iv)
\texttt{ctx\_maxAllowedVelocity} (real value in the range 100-600 with precision
0.1 and mm/s). The selection of the coffee machine is described by four
adaptation rules. Basically, the first two rules prioritize the \texttt{energyConsumption}
when the battery is low (less than 15\%), selecting the nearest
coffee machine. The last two rules prioritize the \texttt{performance} if the battery
is high enough, selecting the machine with lower waiting time (thereby taking
longer travel distance into account but minimizing the overall processing time).

\lstset{ %
  basicstyle=\tiny\ttfamily,           
  numbers=none,                   
  numberstyle=\tiny\color{gray},  
  stepnumber=1,                   
  mathescape=false,
  numbersep=5pt,                  
  backgroundcolor=\color{white},      
  showspaces=false,               
  showstringspaces=false,         
  showtabs=false,                 
  frame=single,                   
  rulecolor=\color{black},        
  tabsize=2,                      
  captionpos=b,                   
  breaklines=true,                
  breakatwhitespace=false,        
  title=\lstname,                   
  keywordstyle=\color{black}\bfseries,          
  commentstyle=\color{darkgreen},       
  stringstyle=\color{blue},         
  morestring=[b]\",
  morecomment=[s]{/*}{*/},
  morekeywords={number,enum,context,var,rule,varpoint,range,precision,unit}               
}

\begin{lstlisting}[caption={VML Model for choosing Coffee Machine}, label=lst:vml-coffee]
/* Data type definitions */
number batteryType { range: [5,100]; precision: 1; }
number velocityType { range: [100,600]; precision: 0.1; unit: "mm/s"; }
number distanceType { range: [0,20]; precision: 0.1; unit: "m"; }
number timeType{range: [10,300]; precision: 1; unit: "s";}
enum machineType { COFFEE_MACHINE_A; COFFEE_MACHINE_B; } 
/* Contexts */
context ctx_battery : batteryLevelType;
context ctx_distanceMachine_A : distanceType;
context ctx_distanceMachine_B : distanceType;
context ctx_waitingTimeMachine_A : timeType;
context ctx_waitingTimeMachine_B : timeType;
context ctx_maxAlowedVelocity : velocityType;
/* Auxiliary variables */
var timeMachine_A = ctx_waitingTimeMachine_A + ctx_distanceMachine_A / ctx_maxAlowedVelocity;
var timeMachine_B = ctx_waitingTimeMachine_B + ctx_distanceMachine_B / ctx_maxAlowedVelocity;
/* Adaptation rules */
rule lowBattery_NearMachineA : ctx_battery < 15 & ctx_distanceMachine_A < ctx_distanceMachine_B => coffeeMachine = COFFEE_MACHINE_A;
rule lowBattery_NearMachineB : ctx_battery < 15 & ctx_distanceCM_A >= ctx_distanceCM_B => coffeeMachine = COFFEE_MACHINE_B;
rule high_EFF_coffeeMachA : ctx_battery >= 15 & timeMachine_A  > timeMachine_B => coffeeMachine = COFFEE_MACHINE_A;
rule high_EFF_coffeeMachB : ctx_battery >= 15 & timeMachine_A <= timeMachine_B => coffeeMachine = COFFEE_MACHINE_B;
/* Variation Points */
varpoint coffeeMachine : machineType;
\end{lstlisting} 

\lstset{ %
  basicstyle=\tiny\ttfamily,           
  numbers=none,                   
  numberstyle=\tiny\color{gray},  
  stepnumber=1,                   
  mathescape=false,
  numbersep=5pt,                  
  backgroundcolor=\color{white},      
  showspaces=false,               
  showstringspaces=false,         
  showtabs=false,                 
  frame=single,                   
  rulecolor=\color{black},        
  tabsize=2,                      
  captionpos=b,                   
  breaklines=true,                
  breakatwhitespace=false,        
  title=\lstname,                   
  keywordstyle=\color{black}\bfseries,          
  commentstyle=\color{darkgreen},       
  stringstyle=\color{blue},         
  morestring=[b]\",
  morecomment=[s]{/*}{*/},
  morekeywords={number,enum,context,var,rule,property,varpoint,range,precision,unit,maximized,minimized,priorities,definitions}               
}

\begin{lstlisting}[caption={VML Model for adapting Velocity and Speaker Volume}, label=lst:vml-velocity]
/* Data type definitions */
number percentType { range: [0,100]; precision: 1; }
number velocityType { range: [100,600]; precision: 0.1; unit: "mm/s"; }
/* Contexts */
context ctx_battery : percentType;
context ctx_noise   : percentType;
/* Adaptation rules */
rule low_noise: ctx_noise < 20 => speakerVolume = 35;
rule medium_noise: ctx_noise >= 20 & ctx_noise < 70 => speakerVolume = 55;
rule high_noise: ctx_noise >= 70 => speakerVolume = 85;
/* Properties */
property performance : percentType maximized {
  priorities: f(ctx_battery) = max(exp(-ctx_battery/15)) - exp(-ctx_battery/15);
  definitions: f(maximumVelocity) = maximumVelocity; }
property energyConsumption : percentType minimized {
  priorities: f(ctx_battery) = exp(-1 * ctx_battery / 15);
  definitions: f(maximumVelocity) = exp(maximumVelocity / 150); }
/* Variation points */
varpoint maximumVelocity : velocityType; 
varpoint speakerVolume : percentType;
\end{lstlisting}

Regarding the second model, shown in listing \ref{lst:vml-velocity}, the
context variables are: (i) the battery level (integer value in the range 0-100)
and (ii) the ambient noise level (integer value in the range 0-100).
In this case, the speaker volume is modified depending on the ambient noise
level as shown in the adaptation rules. 
Furthermore, maximum velocity is adjusted based on the optimization of the 
properties \texttt{performance} and \texttt{energyConsumption}. Note that the
variation point \texttt{maximumVelocity} from this model is used as input (e.g.,
context) in the first model in listing \ref{lst:vml-coffee}.
Finally, we also comment that all the mathematical 
descriptions of the functions in the VML models have been obtained empirically. 

Listing \ref{lst:minizinc} shows the \emph{MiniZinc} model obtained from the VML
model shown in listing \ref{lst:vml-velocity}. The translation from the VML to the
\emph{MiniZinc} models is based on the following mapping rules:
(i) Context variables are translated into parameters (lines 2-3); \mbox{(ii)
Variation} points into decision variables (lines 12-13); \mbox{(iii) Adaptation}
rules and variation points dependencies appear as constraints (lines 15-17); and
(iv) Properties form the cost function to be minimized (lines
33-34), which uses a set of auxillary constraints (lines 19-31)
for linear approximations (since many constraint solvers do not support real
functions).
Note that the constraint solver needs concrete values for the
parameters in the model (the concrete context).
These values can be either fixed in the model or passed to the solver as inputs.

\lstset{ %
  basicstyle=\tiny\ttfamily,           
  numbers=left,                   
  numberstyle=\tiny\color{gray},  
  stepnumber=1,                   
  mathescape=false,
  numbersep=5pt,                  
  backgroundcolor=\color{white},      
  showspaces=false,               
  showstringspaces=false,         
  showtabs=false,                 
  frame=single,                   
  rulecolor=\color{black},        
  tabsize=2,                      
  captionpos=b,                   
  breaklines=true,                
  breakatwhitespace=false,        
  title=\lstname,                   
  keywordstyle=\color{black}\bfseries,          
  commentstyle=\color{darkgreen},       
  stringstyle=\color{blue},         
  morestring=[b]\",
  morecomment=[l]{\%},
  morekeywords={int,float,exp,int2float,constraint,solve,minimize}               
}


\begin{lstlisting}[caption={MiniZinc model obtained from the VML model in
listing \ref{lst:vml-velocity}}, label=lst:minizinc]
% Context Parameters
int: 	ctx_battery;
int: 	ctx_ambientNoise;
% Auxiliary Parameters
float: priority_performance = ( exp(-5.0/15.0) - 
 exp(-1.0 * int2float(ctx_battery) / 15.0) ) / 
 ( exp(-5.0/15.0) - exp(-100.0/15.0) );
float: priority_energy = 
(exp(-1.0*int2float(ctx_battery)/15.0) - exp(-100.0/15.0)) 
 / ( exp(-5.0/15.0) - exp(-100.0/15.0) );
% Variantion points
var 100.0..600.0: maxVelocity;
var 0..100: speakerVolume;
% Constraints
constraint ctx_ambientNoise < 20 -> speakerVolume = 35;
constraint ctx_ambientNoise >= 20 /\ ctx_ambientNoise < 70 -> speakerVolume = 55;
constraint ctx_ambientNoise >= 70 -> speakerVolume = 85;

var 0.0..100.0: aux;
constraint maxVelocity <= 100.0 
   -> aux = 1.768/100.0 * maxVelocity;
constraint maxVelocity > 100.0 /\ maxVelocity <= 200.0	
   -> aux = 3.444/100.0 * maxVelocity - 1.676;
constraint maxVelocity > 200.0 /\ maxVelocity <= 300.0
   -> aux = 6.708/100.0 * maxVelocity - 8.204;
constraint maxVelocity > 300.0 /\ maxVelocity <= 400.0
   -> aux = 13.07/100.0 * maxVelocity - 27.29;
constraint maxVelocity > 400.0 /\ maxVelocity <= 500.0
   -> aux = 25.44/100.0 * maxVelocity - 76.77;
constraint maxVelocity > 500.0 /\ maxVelocity <= 600.0
   -> aux = 49.57/100.0 * maxVelocity - 197.42;
% Solver mode
solve minimize priority_performance * 
 ( -1.0*100.0 * (maxVelocity-100.0) / (600.0 - 100.0) ) + priority_energy * aux;
\end{lstlisting}

%
%

In order to validate our approach in a system of realistic
complexity, we integrated the coffee delivery example into our Butler scenario
(see our YouTube channel\footnote{\url{http://www.youtube.com/user/roboticsathsulm}}).

At this point we have all the ingredients we need to put everything into
operation, i.e., the VML models that specify the design-time open variation
points and the mechanisms that allow the robot to bind them at run-time.
Therefore we use a component with a constraint solver that interprets MiniZinc
models. A noteworthy issue is the interaction of the constraint solver in our
architecture. As input, this component needs context information which is
typically distributed in the system on different levels. In \textsc{SmartSoft}
we use an event based mechanism to acquire certain information from skill
components. For instance, for both models it is necessary to get the battery
value.
Thereto the constraint solver component directly subscribes to the base
component to be informed if the battery value drops below a certain threshold.
More advanced data acquisition mechanisms (e.g., those needed to collect
information to deduce the waiting time in front of a coffee machine) can be implemented 
based on a data aggregation approach like in~\cite{Lotz:2011}.
Finally, some context information is related to the current
environment model of the robot, that is stored and updated in the knowledge base
of the sequencer (see top right in fig.~\ref{fig:chs-architecture}). As the
sequencer is always the master in our system it has to provide the relevant
information to the constraint solver. For example the context variable 
\texttt{ctx\_maxAllowedVelocity} in list.~\ref{lst:vml-coffee} is stored in
the knowledge base, because it relates to the current physical limits of the robot, the
current situation, etc.

The next question relates to the triggering of the adaptation mechanism. In our
system we separate the triggering (execution) of the constraint solver from its
model implementation. This enables the VML modeller to focus on adaptation
without having to care about the execution environment.
There are two possible situations where the constraint solver is triggered to
calculate new values for certain variation points. In the first situation the
sequencer queries the constraint solver -- providing current context information
in the query request -- for a decision to e.g., expand the current node in its
task-net (see fig.~\ref{fig:chs-architecture} (c)). In this case the sequencer
triggers the calculation in the constraint solver and waits for the (query)
result. In the second situation a variation point (e.g., the \texttt{speakerVolume} in
listing~\ref{lst:vml-velocity}) is sporadically updated on demand each time the
constraint solver receives an event about changes in relevant context variables.
Again, using the event mechanism here considerably reduces communication
and calculation overhead in the system. The \texttt{speakerVolume} variation
point can be directly parametrized in the speech output component, because it typically has
no effect on the sequencer. As mentioned above, other variation points like the
currently suggested speed reduction must be additionally propagated to the
sequencer in order to guarantee a consistent world model in the knowledge base.

Figure~\ref{fig:scenario} illustrates the behavior of our robot Kate according
to the VML model in listing~\ref{lst:vml-coffee}. Thereby, our robot Kate
navigates in our home environment and decides, for each order on demand, which
coffee machine to use. For this purpose, Kate balances between conflicting
optimization strategies like minimizing resource consumption while keeping high
the task efficiency.

\begin{figure}[htpb]
   \centering
   \includegraphics[width=1.0\columnwidth]{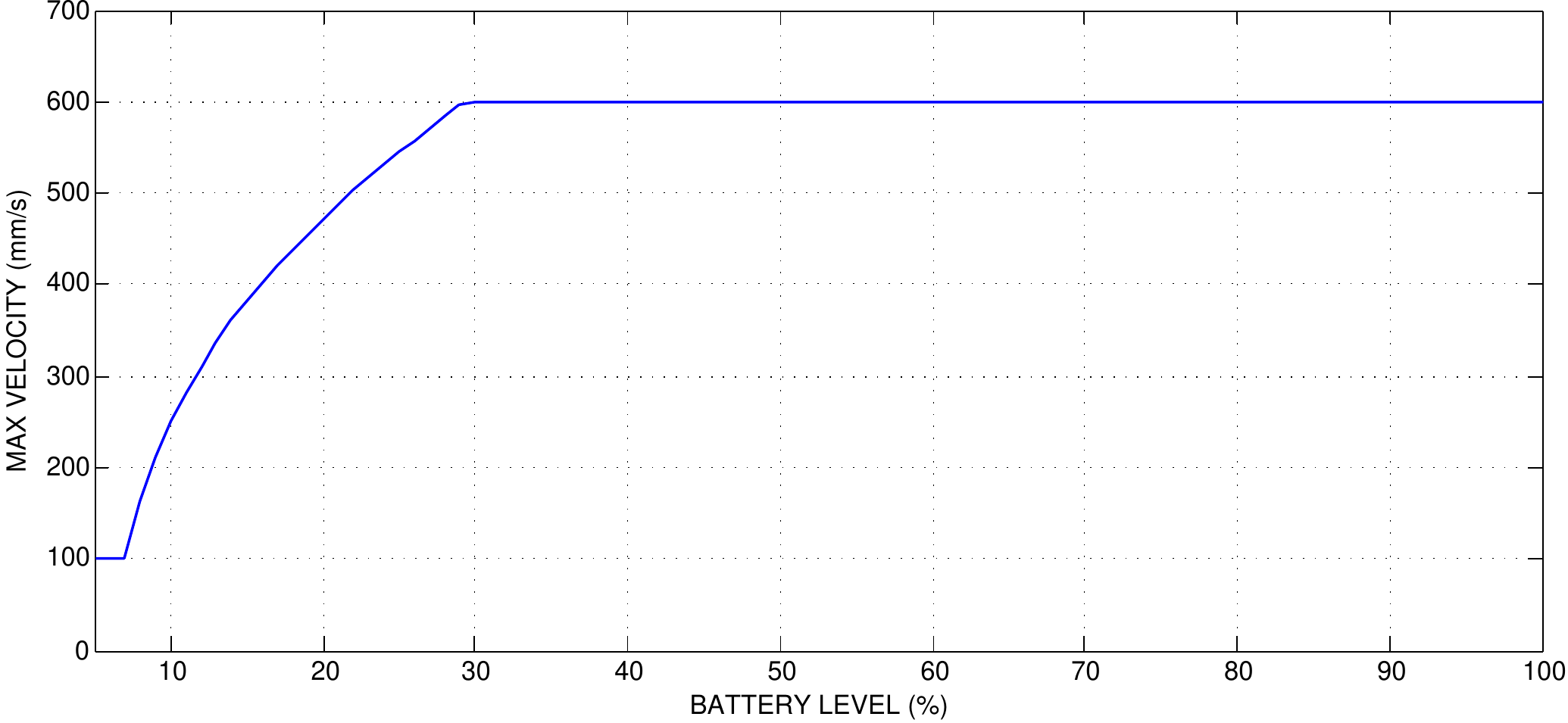}
   \caption{Evolution of maximum allowed velocity according to the current battery level. Results after running the VML Model in listing
   \ref{lst:vml-velocity}.}
   \label{fig:scenario2}
\end{figure}

Figure 5 shows the evolution of the variation point maximum allowed 
velocity by balancing \texttt{performance} and \texttt{energy\-Consumption}
according to the current battery level. While the battery level is higher than 30\% the 
selected maximum velocity remains at 600 mm/s, since \texttt{performance} has priority 
over \texttt{energy\-Consumption}. However, if the battery falls below 30\%, 
\texttt{energy\-Consumption} becomes increasingly important, therefore the
selected maximum velocity decreases exponentially until reaching the minimum of 100 mm/s.

\section{Related Work}

%
%

\subsection{Variability in Robotics}

In the last decade, the robotics community has spent a lot of efforts 
to improve the robotics software development process in order to cope with its
inherent and growing complexity and to improve its reusability, 
interoperability and maintainability. Many of these efforts have adopted Software
Engineering best practices like those promoted by CBSE and MDE. 
As a result, a new set of robotics frameworks and architecture models, aimed to
improve robotics software design, have been developed, like \emph{RT-Component}~\cite{Ando:2005}, 
\emph{ROS}~\cite{Quigley:2009}, \textsc{SmartSoft}~\cite{Schlegel:2010}, etc.

Although all these approaches are already valuable steps towards building
impressive systems, we believe, that improving only the development process at
design-time in isolation is insufficient, and it is rather necessary to take
run-time aspects like situation- and resource-awareness into account from the
very beginning. One step to solve this problem is done by introducing
mechanisms to model robot tasks independently of the reactive components, e.g.,
addressed by \emph{SMACH}~\cite{Boren:2010} and
\textsc{SmartTCL}~\cite{Steck:2011,Steck:2011:GPCE}.
The latter approach additionally allows to easily model dynamic task trees, that
are rearranged on demand according to current situation and other parameters
like resource consumption or adaptation suggestions (see sec.
\ref{sec:integration-into-robotics}). Thus, one of the results
in this paper is to explicate properties which are necessary to be expressed at
design-time by a developer and which can be later processed on the robot at
run-time.

In a robotic system, variability can be identified on different levels. State-of-the-art systems focus mostly on functional variability like in
\cite{Gherardi:2011} (based on feature models for components). By contrast, our approach is not limited to design-time variability. Instead,
we introduce a DSL tailored to express variation points at design-time and their binding according to non-functional properties at run-time.

\emph{DiaSpec}~\cite{Cassou:2011} is a recently introduced Java based design language to model robotic systems on the skill layer. In addition
thereto, our approach includes the sequencing layer and is more focused on variability modeling.

The software community introduced the term \emph{architecture-based adaptation} in a series of papers like \cite{Georgas:2008} and \cite{Edwards:2009} 
for robotic systems based on CBSE. In there, components are replaced or migrated at run-time (e.g., due to a failure or resource insufficiency) by
similar components with differently implemented services. From the robotics perspective we highly support this progress.

However, in contrast, the focus in our work is more on the 
challenge to balance between reduction of complexity at design-time and at the
same time introducing models to express variability for non-functional
properties which can be efficiently exploited at run-time. Thereby, we use
sophisticated real-world scenarios from service-robotics to demonstrate the
applicability of our approaches in robotic systems of realistic complexity.

%
%


\subsection{Expressing run-time variability in Software Engineering}

Numerous research works in software engineering have investigated how to model
the adaptation logic~\cite{Salehie:2009}. Therefore, we can find a wide range
of approaches that address, among others, different representations and
formalisms, different levels of abstraction, different application domains, and
different techniques to capture and express adaptation. Regarding approaches
using different representation for expressing adaptation, in the area of
software architecture, we can find \emph{Architectural Description Languages
(ADL)}~\cite{Bradbury:2004} based on graphs, process algebras, and
other formalisms to describe desired component configurations and specify how
configurations may be changed at run-time in terms of addition and removal of
components. In contrast, in the area of \emph{Dynamic Software Product
Lines}~\cite{Fleurey:2009,Cetina:2009}, the system variability is modelled
using variation points defined with a number of alternatives and constraints
among them. Concerning the level of abstraction, while \cite{Fleurey:2009}
provides designers with a high level modeling language to specify the global
adaptation of the system using fuzzy logic and qualified variables (e.g., the
QoS properties used to decide adaptation are expressed as enumerators defined
with values like HIGH, MEDIUM or LOW), \cite{Hallsteinsen:2006} allows a wide
range of data types and utility functions to specify QoS properties in the
components of the system. Moreover, although most of the approaches are
independent of the application domain, some are focused, e.g., on the dynamic 
adaptation of Graphical User Interfaces using model transformations
\cite{Criado:2012}, or in the use of feature diagrams to describe the 
functional variability of Smart Houses~\cite{Cetina:2009}.
Finally, a number of
techniques have been proposed in the literature to capture and express
adaptation logic. First, most existing approaches are based on using \emph{ECA
(event-condition-action)} rules
\cite{Criado:2012,Cetina:2009,Hussein:2011}. In these approaches the
context and the configurations are related by a set of rules, which express how
the changes of the system context should affect the running configuration of
the application. ECA rules are clear and easy to write but fully specifying an
adaptive system using ECA rules often requires defining a large set of rules
\cite{Fleurey:2009}. In addition, it is not easy to check whether the full set 
of rules is consistent or not at design-time and it might be the case that
some rules conflict with others, in particular when some changes in the context 
might trigger several rules. Another common way to express the adaptation
logic is to define goals that the system should reach \cite{Hallsteinsen:2006}.
The designer establishes QoS properties and specifies the impact on them
depending on the selected configuration, e.g., through utility
functions. At run-time, the system should find the best configuration i.e.,
the one that optimizes the properties considering the current context. Goals allow
to specify the adaptation logic at a higher level of abstraction than ECA
rules. However, multi-dimensional optimization algorithms are usually resource
and/or time-consuming. 
Some approaches, like the one presented in~\cite{Fleurey:2009}, adopt a combined 
approach based on the use of both rule- and optimization-based mechanisms, to try
exploiting their respective advantages while limiting their drawbacks.

Our approach is close to the one presented in~\cite{Fleurey:2009} but, in contrast, we do not rely 
on fuzzy logic to capture and describe how systems should adapt. Conversely, we offer 
a more precise and less limited way to describe variability, e.g., using mathematical 
expressions that incorporate any number of real variables. This provides designers 
with a more natural way for describing the variability of their systems, 
in particular in some application domains like in robotics.

Our proposal is also close to the one presented in \cite{Hallsteinsen:2006}, 
although it relies on component-based system adaptation while ours is independent 
of the system organization (variation points can be components, algorithms,
parameters, etc.).
Therefore, the proposed modeling language can describe variability at
different levels of abstraction. Although we
focus on the robotic domain, all the concepts included in our approach are
application independent like \cite{Fleurey:2009,Hallsteinsen:2006} or
\cite{Hussein:2011}. Finally, as in \cite{Fleurey:2009}, we have considered ECA
rules and optimization of goals, but with some improvements like unit
specification and normalization.

\section{Conclusions and Future Work}

%
%

In this paper we showed how to express variability in a robotic system for 
non-functional properties (like safety or task efficiency), using 
a first version of a new DSL called VML. This language enables designers to
focus on modeling the adaptation strategies without having to foresee and explicitly 
deal with all the potential situations that may arise in real-world and
open-ended environments. The variability, purposefully left open by the designers 
in the VML models, is then fixed by the robot at run-time according to
its current tasks and context (separation of roles and concerns). Furthermore,
we underpinned the applicability of our approach by integrating it into our
overall robotic architecture and by implementing it in a sophisticated
real-world scenario on our service robot Kate. Thereto the VML models
were translated into a constraint modeling language (MiniZinc), which is
executed in a constraint solver interacting with the sequencing layer on the
robot at run-time.

Considering that a VML model represents a constrained optimization problem, the
question arises why VML is used instead of a modeling language for constraint
programming. This discussion resembles the traditional dispute between
\emph{Domain-Specific Languages (DSLs)} and \emph{General-Purpose Languages
(GPLs)}. VML has the well-known advantages of DSLs (e.g., allowing solutions to
be expressed at the level of abstraction of the problem domain considerably
simplifying the work of the designers) and disadvantages (e.g., cost of
designing, implementing and maintaining a new language) \cite{Mernik:2005}.
However, we highlight that the proposed language is of value because: (i) it
enables designers to clearly express an important issue, i.e., the run-time
variability for optimizing the execution quality of the system; (ii) it delimits the
concepts and their semantics which facilitates the validation of the models;
(iii) it abstracts some details such as the normalization of the formulas; and
(iv) it can be extended adding new concepts and capabilities not supported by
the expressiveness of the constraint modeling languages.

As a conclusion, we were able to combine efforts from the different communities 
(SE, MDE and Robotics) in order to apply state-of-the art approaches for
variability management on a robot operating in a home-like environment.

For the future, we plan to extend VML with some additional syntax constructs 
and to improve the supporting tools, in particular the VML model editor, 
to provide designers with some advanced model validation and simulation facilities.

\section{Acknowledgments}
\small{This collaboration is jointly funded by the German Academic Exchange
Service (DAAD; Project ID: 54365646; Project leader: Prof. Schlegel;
\url{http://www.zafh-servicerobotik.de}) and the General Directorate of
International Projects of the Spanish Ministry of Economy and Competitiveness
(MINECO; Project leader: Dr. Vicente-Chicote; Project ID: PRI-AIBDE-2011-1094).
Juan F. Ingl\'{e}s-Romero thanks Fundaci\'{o}n S\'{e}neca-CARM for a research
grant (Exp. 15561/FPI/10).}

\small{The authors also want to thank Matthias Lutz for his extraordinary
support in integrating the examples on the service robot Kate.}

\addtolength{\textheight}{-12cm}   


\bibliographystyle{IEEEtran}
\bibliography{IEEEabrv,DSLRob-2012}

\end{document}